\pdfoutput=1
\documentclass[11pt]{article}
\usepackage{acl}

\usepackage{times}
\usepackage{latexsym}

\usepackage[LGR, T1]{fontenc}
\usepackage[utf8]{inputenc}
\usepackage{microtype}

\usepackage{etruscan}
\usepackage{textgreek}
\usepackage{alphabeta}
\usepackage{tipa}
\usepackage{amsmath,amssymb,amsfonts}
\usepackage{algorithmic}
\usepackage{graphicx}
\usepackage{textcomp}
\usepackage{xcolor}
\usepackage{realhats}
\usepackage{hyperref}       % page numbers and '\ref's become clickable
\usepackage{todonotes}      % notes
\usepackage{float}          % for forcing images to stay where they should
\usepackage{multirow}       % for using multirow in tables
\usepackage{fancyhdr}       % page numbers
\usepackage{enumitem,cleveref} % ref for enumerate lists
\usepackage{siunitx}        % scientific notation
\usepackage{array}
\usepackage{supertabular}

\newcolumntype{L}[1]{>{\raggedright\let\newline\\\arraybackslash\hspace{0pt}}m{#1}}
\newcolumntype{C}[1]{>{\centering\let\newline\\\arraybackslash\hspace{0pt}}m{#1}}
\newcolumntype{R}[1]{>{\raggedleft\let\newline\\\arraybackslash\hspace{0pt}}m{#1}}

\newcommand{\note}[1]{}

\renewcommand{\th}{\textsuperscript{th}}
\newcommand{\st}{\textsuperscript{st}}

\newcommand{\token}[1]{<#1>}
\newcommand{\mtoken}[1]{\text{\token{#1}}}

\usepackage{tikz}
\usetikzlibrary{automata,positioning,decorations.text,topaths,arrows.meta,decorations.pathmorphing,quotes}

\usepackage[greek, english]{babel}

\title{Larth: Dataset and Machine Translation for Etruscan}
\renewcommand\footnotemark{} % Remove mark from \thanks

\author{Gianluca Vico \and Gerasimos Spanakis\\
  Department of Advanced Computing Sciences / Paul-Henri Spaaklaan 1\\
  Maastricht University / Maastricht, The Netherlands \\
  \texttt{g.vico@student.maastrichtuniversity.nl} \\
  \texttt{jerry.spanakis@maastrichtuniversity.nl}\\}

\begin{document}
\maketitle

\begin{abstract}
% \note{TODO\\
% - topic\\
% - what we did\\
% - main results}
% - Etruscan is an ancient language with few resources
% - to the best of our knowledge there is no a publicly available dataset for nlp
% - we propose a dataset and demonstrate it is suitable for translation

% Acquired partially manually and partially automatic from existing academic sources
% Releasing the dataset and the model can help better understanding the language

Etruscan is an ancient language spoken in Italy from the 7\th century BC to the 1\st century AD. 
There are no native speakers of the language at the present day, and its resources are scarce, as there exist only around 12,000 known inscriptions.
To the best of our knowledge, there are no publicly available Etruscan corpora for natural language processing.
Therefore, we propose a dataset for machine translation from Etruscan to English, which contains 2891 translated examples from existing academic sources.
Some examples are extracted manually, while others are acquired in an automatic way.
% \jnote{add a few more details on how you acquired it etc.}
Along with the dataset, we benchmark different machine translation models observing that it is possible to achieve a BLEU score of 10.1 with a small transformer model. 
%\jnote{add one sentence about how this creates new opportunities as a corpus and a model but also why it can be useful? Also, the goal is to release the dataset right? we can also mention that}
Releasing the dataset \footnote{The data and code are available here: \url{https://github.com/GianlucaVico/Larth-Etruscan-NLP.git}} can help enable future research on this language, similar languages or other languages with scarce resources.
\end{abstract}

% https://github.com/GianlucaVico/Larth-Etruscan-NLP.git

\section{Introduction}

Etruscan (ISO 639-3 code: \texttt{ett}) is a language spoken in the Etruria region (modern-day centre Italy) from the 7\th century BC to the 1\st century AD \citep{zihk_rasna}.
It is written right to left using the Etruscan alphabet, derived from the Greek alphabet \citep{zihk_rasna}.
The predominant word order in this language is mostly subject-object-verb \citep{zihk_rasna}. This pattern is similar to Latin, but distinguishing it from other languages like English, where the words follows the subject-verb-object order. It has 5 cases (accusative, nominative, genitive, dative and locative), two numbers (singular and plural) and takes into consideration animacy and gender \citep{zihk_rasna}.

Only a small number of inscriptions in this language survived up to the present day: an estimated 12,000 inscriptions have been recovered \citep{zihk_rasna}.
However, only a few of them have a significant length to be considered complete. Other ancient languages used in similar areas and periods in history, such as Latin and Ancient Greek, have more resources, thus, making natural language processing techniques and tools easier to develop for these languages.

% Introduce other ancient languages
% What this paper is trying to do
% More details
The contribution of this paper is threefold: First, we build a corpus of Etruscan inscriptions usable for natural language processing.
We use as a starting point existing academic resources for this language exist, and we try to create our corpus both by manual and automatic work. Second, we focus on the machine translation task from Etruscan to English. We evaluate whether neural models can be trained with this data and if they can outperform less data-hungry models. Finally, we investigate if it is possible to exploit any similarity between Etruscan and Latin or Ancient Greek to improve the aforementioned model. 

%\section{Outline}
%\note{Remove section title?}
In Section \ref{sec:sota}, we introduce state-of-the-art techniques relevant to this paper. Then, in Sections \ref{sec:data} and \ref{sec:mt} we explain the methods used to work on the data and the model used. Section \ref{sec:experiments} and Section \ref{sec:results} illustrate the experiments and compare the different techniques. Finally, Section \ref{sec:conclusion} concludes the paper.

\section{Literature review} \label{sec:sota}

% Etruscan data -> manuals, ciep
% An example of a digital Etruscan corpus is the Etruscan Texts Project (ETP) \citep{etp} which contains 369 inscriptions.
The Etruscan Texts Project (ETP) \citep{etp} is a digital Etruscan corpus which contains 369 inscriptions.
The project is based on Etruskische Texte \citep{et} and is used in the book Zihk Rasna \citep{zihk_rasna}.
Another digital Etruscan work is the Corpus Inscripionum Etruscarum Plenissimum (CIEP) \citep{ciep}, based on the Corpus Inscriptionum Etruscarum (CIE) \citep{cie}.

% Latin data and CLTK
Similar works exist for Latin and Ancient Greek, like I.PHI \citep{iphi} and Perseus \citep{perseus}.
In addition, toolkits like CLTK \citep{cltk} offer natural language processing for these languages.
Projects that aim to increase the resources available for low-resource languages may also include ancient languages, like the Tatoeba Translation Challenge \citep{tatoeba}.
It has Latin and Ancient Greek datasets, however, it does not include Etruscan.

% MT, SMT (IBM)
% Translating from Etruscan to English can be solved via neural machine translation \citep{nmt}.
The machine translation task can be solved via neural machine translation \citep{nmt}, which involves training neural networks that take texts from the source language and generate the translation in the target language. 
Popular architectures include Long short-term memory (LSTM) \citep{lstm} and transformers \citep{attention}.
These models are sequence-to-sequence \citep{seq2seq}, meaning they take a sequence as input and generate a sequence of possibly different lengths as output.
One approach is to feed word or word pieces to the model like in T5 \citep{T5} or \citet{NeuralMT}.
\citet{character-nmt1} and \citet{character-nmt2} show that it is possible to work directly on characters, while other models (\citealp{char-word1} and \citealp{char-word2}) use a hybrid approach by working on both the character and word sequences.

Besides neural networks, other approaches include rule-based models, such as dictionary models, which translate the text based on explicit rules, and statistical models \citep{smt}. %\todo{Add more}

By using the transformer architecture, Ithaca \citep{ithaca} is able to perform textual restoration and geographical and chronological attribution of ancient Greek inscriptions.
The model consists of a sparse self-attention encoder \citep{bigbird} that takes as input the characters and the words of the input text, and then three feed-forward blocks generate the output for each task.
Other examples of transformer models working on ancient languages are the multi-language translation model Opus-MT \citep{opus}, 
%which achieves BLEU 21.5 and chr-F 0.402 
tested on the Latin $\rightarrow$ English split of the Tatoeba dataset, or the language model Latin-BERT \citep{latin-bert}.

Translation models can be evaluated by using various metrics. 
\citet{bleu} proposes BLEU: this metric considers the average matching precision of n-grams between the reference text and the machine-translated text.
% A variation of BLEU is NIST, while METEOR also addresses the alignment problem between the source and target languages.
Another metric is TER \citep{ter}, which measures the quality of the translation based on the number of edits needed to change the system text to the reference one.
TER and BLEU are based on word n-grams, while chr-F \citep{chrf} uses the F-score of matching character n-grams.

% \section{Methodology} \label{sec:methodology}

\section{Data} \label{sec:data}
\subsection{Etruscan}
First, we collect a dataset containing Etruscan texts. 
The main sources used are CIEP \citep{ciep}, ETP \citep{etp}, and the book "Zikh Rasna: A Manual of the Etruscan Language and Inscriptions" \citep{zihk_rasna}, which cites "Etruskische Texte" \citep{et}.
It is possible to extract Etruscan inscriptions and their translations where available from ETP and Zikh Rasna. 
In addition, we extract the date and location of the inscriptions.
Also, Zikh Rasna contains a list of Etruscan words and proper names used to make a dictionary.
From CIEP, we extract only the inscriptions and the translations. 
However, the inscriptions are often incomplete or noisy due to the structure of CIEP itself and the limitation of the PDF extracting software (PyMuPDF, \citealp{pymupdf}).
% Moreover, Zikh Rasna contains a list of Etruscan words and proper names used to make a dictionary. % \todo{Number of words and translations}
We make two datasets. The first, \textbf{ETP}, uses data from ETP and ZIkh Rasna, while the second \textbf{ETP+CIEP}, adds the data from CIEP.

After removing strings that are in the wrong language, the text is normalised.
CIEP and ETP use two different transcription conventions. 
Also, Etruscan uses several symbols as word separators (" ", "$\cdot$", ":", "\vdots"), which are converted to white space (" ").
Table \ref{tab:transcription} illustrates how the Etruscan alphabet is transcribed by ETP and by us (Larth).
Note that the transcription is not reversible.

\begin{table}[t]
    \centering
    \begin{tabular}{c|c|c}
    % \hline
    \textbf{Etruscan} & \textbf{ETP} & \textbf{Larth}\\
    \hline%\hline
    \textetr{\Aalpha} & a & a \\
    \textetr{\Abeta(*)} & b & b \\
    \textsf{C} & c & c \\
    \textetr{\Adelta(*)} & d & d \\
    \textetr{\Aepsilon} & e & e \\
    \textetr{\Adigamma} & v & v \\
    \textetr{\Azeta} & z & z \\
    \textetr{\Aeta} & h & h \\
    \textetr{\Atheta} & \texttheta & th \\ 
    \textetr{\Aiota} & i & i \\
    \textetr{\Akappa} & k & k \\
    \textetr{\Alambda} & l & l \\
    \textetr{\Amu} & m & m \\
    \textetr{\Anu} & n & n \\
    $\boxplus$ & $\overset{\text{\tiny +}}{\text{s}}$ & s \\
    \textetr{\Aomicron} & o & o \\
    \textetr{\Aesade} & \textsigma, {\'\textsigma} & s, sh\\
    \textetr{\Api} & p & p \\
    \textetr{\Aqoph} & q & q \\
    \textetr{\Arho} & r & r \\
    \textetr{\Asigma} & s, {\'s}, \textvarsigma, \'\textvarsigma & s, sh, s, sh \\
    \textetr{\Atau} & t & t \\
    \textsf{V} & u & u \\
    \textetr{\Achi} & \textovercross{s} & sh \\ % $\overset{\textsf{\tiny x}}{\text{s}}$
    \textetr{\Aphi} & \textphi & ph \\
    \textetr{\Apsi} & \textchi & kh \\
    \textetr{\Avau} & f & f \\
    % \hline
    \end{tabular}
    \caption{Texts from ETP are already transliterated, but CIEP transliteration is sometimes ambiguous. We further reduce the number of symbols by using a subset of the Latin alphabet.
    % (*): these letters are in the alphabet but Etruscan does not use them.
    }
    \label{tab:transcription}
\end{table}

In the end, we obtain 7139 Etruscan texts (561 from ETP and 6578 from CIEP). Among these, a translation is available for only 2891 inscriptions (239 from ETP and 2652 from CIEP).
Also, the vocabulary built from ETP contains 1122 words, of which 956 with a translation.
Each word is also described by 54 binary grammatical features (e.g., plural, active, passive, ...).
The type of text is not included in the dataset, however, ETP lists on their website mostly proprietary and funerary texts \cite{etp} (137 and 104 out of 369).

\begin{figure}[t]
    \centering    
    \includegraphics[width=0.8\columnwidth]{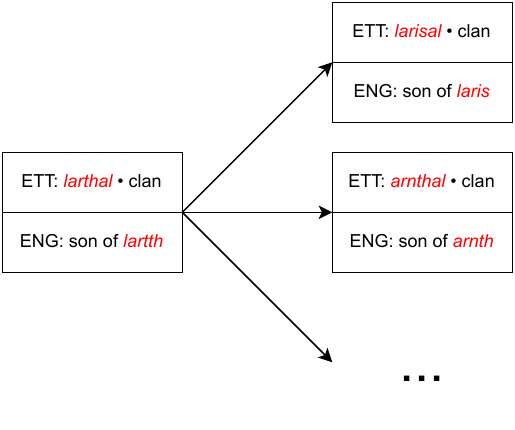}
    \caption{Example of data augmentation by replacing proper names. The name is replaced both in the Etruscan text and the English translation.}
    \label{fig:name-augm}
\end{figure}

Since the data is limited, we perform data augmentation.
Many inscriptions contain proper nouns, so we use the dictionary we built to replace them with other proper nouns with the same grammatical features. 
The substitution is done simultaneously on the Etruscan and English texts in order to keep the translations correct, as shown in Figure \ref{fig:name-augm}.
Also, inscriptions can be damaged, so parts of the words cannot be read and the translation models have to either discard those words or rely only on the remaining characters.
So, we generate more training samples by damaging more words.
We assume that the damage occurs at the beginning or end of the words with a set probability.
Also, we assume the number of damaged characters follows a geometric distribution.
In this way, for instance, the word "clan" can stay unchanged or it might become "--an", "cla-", "-l--".

\subsection{Latin and Ancient Greek}

Models introduced later in the paper use Latin or Ancient Greek documents.
Tatoeba eng-lat \citep{tatoeba} is used to train the Latin model. 
The text is normalised and non-Latin characters are removed.
For Ancient Greek, we use Perseus \citep{perseus}. 
In this case, we also remove all diacritical marks and transliterate the text to Latin. 
In this way, all the languages used share the same alphabet.

\section{Machine Translation} \label{sec:mt}
% The main task is to translate from Etruscan to English. 
% We compare different models. 
% We compare different models for machine translation.
% First, we use a random model as a baseline model,
% Then, we test a dictionary-based model and statistical models.
% Finally, we try neural models. 
% The training is  directly on the Etruscan $\rightarrow$ English task, but for the neural networks, we also try to adapt models trained previously on the Latin $\rightarrow$ English and Ancient Greek $\rightarrow$ English to Etruscan.
We compare different models for machine translation on the BLEU metric but chr-F and TER metrics are also reported. 
The metrics are computed by SacreBLEU \citep{sacrebleu}.
Higher BLEU and chr-F and lower TER indicate a better-performing model.
Moreover, we evaluate the case where we use only ETP and ETP+CIEP for training and testing the models.

\subsection{Random Model}

% A random model is used as a baseline model for translating Etruscan $\rightarrow$ English. 
The output of this model does not depend on the Etruscan inputs, but only on the training translations.
It assumes that the length of the translations follows a normal distribution whose parameters are estimated from the training data. 
% In order to avoid negative lengths, it takes the absolute value.
Then, it samples English tokens from the training distribution.
% This model is trained on random training sets from ETP only, CIEP only and both. 
% The remaining data is used for testing.
The experiment is repeated 10 times with random splits of the dataset in training and testing data.
The resulting metrics are then averaged.

\subsection{Dictionary-based Model}

The second model is a dictionary-based model based on the vocabulary provided in Zihk Rasna \citep{zihk_rasna}.
The model assumes that each word has one meaning and one translation.
Moreover, it does not rearrange the word order and it does not consider the grammar of the source language or the target language.
This model splits the input text into word tokens. Then, for each token, it searches for the exact match in the dictionary.
If a match is found, it adds the translation to the output; otherwise, the token is ignored.

\subsection{N-gram and Naïve Bayes Models}

Then, we try to translate Etruscan taking into consideration the previous $n$ tokens. 
The model estimates the probability distribution $\mathbb{P}(eng_i | ett_i, ett_{i-1}, ... ett_{i-n})$, where $eng_i$ and $ett_i$ are tokens at position $i$. 
This is done either directly from the training data or as a Na\"ive Bayes model with the following expression:
%$\mathbb{P} (eng_i | ett_i, ett_{i-1}, ... ett_{i-n}) \propto \mathbb{P}(eng_i) \prod_{j=0}^{n} \mathbb{P}(ett_{i-j} |eng_i)$.
\begin{multline}
\mathbb{P} (eng_i | ett_i, ett_{i-1}, ... ett_{i-n}) \propto \\  \propto \mathbb{P}(eng_i) \prod_{j=0}^{n} \mathbb{P}(ett_{i-j} |eng_i)
\end{multline}

% The translation is generated greedily: the model scans through the input and generates the most probable token. 

% The model assumes that the English and the Etruscan sequences are aligned and that one Etruscan token is translated into one English token. 
The model assumes that one $n$\th Etruscan token is translated into the single $n$\th English token.
Figure \ref{fig:n-gram-mono} shows how the sequences are aligned and which Etruscan context is used for each English token.

A second N-gram model also includes the previous English tokens in the context by computing $\mathbb{P}(eng_i | ett_i, ... ett_{i-n}, eng_{i-1}, ..., eng_{i-n-1})$ as shown in Figure \ref{fig:n-gram-bi}.
% For both approaches, we consider the case when the word order is taken into account and when it is not.
When the probability distribution is estimated directly, we consider the case when the word order is taken into account and when it is not.
We use beam search to generate the output.

\begin{figure}[t]
    \centering

    \begin{tikzpicture}[
        auto, 
        state/.style={circle, draw, minimum size=4em, fill={rgb:black,1;white,10}, font=\small\sffamily},
        node distance = 2mm and 1mm,
    ]
    \node[state] (eng1) {$<$pad$>$};
    \node[state] (eng2) [right=of eng1] {$<$pad$>$};
    \node[state] (eng3) [right=of eng2] {$<$eng1$>$};
    \node[state] (eng4) [right=of eng3] {$<$eng2$>$};
    
    \node[state] (et1) [below=of eng1] {$<$pad$>$};
    \node[state] (et2) [right=of et1] {$<$pad$>$};
    \node[state] (et3) [right=of et2] {$<$ett1$>$};
    \node[state] (et4) [right=of et3] {$<$ett2$>$};

    \node[] (out) [left=of eng1]{Eng.:};
    \node[] (in) [left=of et1] {Ett.:};

    \path[->] 
    (et1) edge (eng3)
    (et2) edge (eng3)
    (et3) edge (eng3);
    \path[dashed, ->] 
    (et2) edge (eng4)
    (et3) edge (eng4)
    (et4) edge (eng4);
    
    \end{tikzpicture}
    \caption{The first approach for the n-gram model. \textit{\token{eng n}} indicates English tokens, while \textit{\token{et n}} are Etruscan tokens; \textit{\token{pad}} is the padding token. The example shows $P({\mtoken{eng1}}|{\mtoken{pad}\mtoken{pad}\mtoken{et1}})$ and $P({\mtoken{eng2}}|{\mtoken{pad}\mtoken{et1}\mtoken{et2}})$. The context is made up of Etruscan trigrams.}
    \label{fig:n-gram-mono}
\end{figure}

\begin{figure}[t]
    \centering

    \begin{tikzpicture}[
        auto, 
        state/.style={circle, draw, minimum size=4em, fill={rgb:black,1;white,10}, font=\small\sffamily},
        node distance = 2mm and 1mm,
    ]
    \node[state] (eng1) {$<$pad$>$};
    \node[state] (eng2) [right=of eng1] {$<$pad$>$};
    \node[state] (eng3) [right=of eng2] {$<$eng1$>$};
    \node[state] (eng4) [right=of eng3] {$<$eng2$>$};
    
    \node[state] (et1) [below=of eng1] {$<$pad$>$};
    \node[state] (et2) [right=of et1] {$<$pad$>$};
    \node[state] (et3) [right=of et2] {$<$ett1$>$};
    \node[state] (et4) [right=of et3] {$<$ett2$>$};

    \node[] (out) [left=of eng1]{Eng.:};
    \node[] (in) [left=of et1] {Ett.:};

    \path[->] 
    (et2) edge (eng3)
    (et3) edge (eng3)
    (eng1) edge [bend left=45](eng3)
    (eng2) edge [bend left=45](eng3);
    \path[dashed, ->] 
    (et3) edge (eng4)
    (et4) edge (eng4)
    (eng2) edge [bend left=45] (eng4)
    (eng3) edge [bend left=45] (eng4);
    
    \end{tikzpicture}
    \caption{the second approach for the n-gram model. \textit{\token{eng n}} indicates English tokens, while \textit{\token{ett n}} are Etruscan tokens; \textit{\token{pad}} is the padding token. The example shows $P({\mtoken{eng1}}|{\mtoken{pad}\mtoken{ett1}}, {\mtoken{pad}\mtoken{pad}})$ and $P({\mtoken{eng2}}|{\mtoken{ett1}\mtoken{ett2}}, {\mtoken{pad}\mtoken{eng1}})$. The context is made up of Etruscan and English bigrams.}
    \label{fig:n-gram-bi}
\end{figure}

\subsection{IBM Models}

Next, we compare our models to existing ones. 
To do so, we consider the IBM models \citep{smt} from the NLTK package \citep{nltk}.
They are a series of 5 models with increasing complexity.
These models consider the alignment between the source strings and the target strings, however, the Etruscan-English pairs we are using do not contain this information.
Therefore, we test the models as if the sequences were aligned.

IBM1 does not consider the word order. 
IBM2 introduces the word order, while IBM3 takes also into consideration that a word can be translated into zero or more words.
IBM4 and IBM5 can also reorder the output words.
Moreover, IBM4 and IBM5 also need the part-of-speech (POS) tags of both the source and target sequences. POS tags are inferred from the grammatical features listed in the dictionary.
For Etruscan, these are obtained by a manually annotated list of words, while the English sequences are tagged by NLTK perceptron tagger.

\subsection{Transformer Models - Larth}

\begin{figure}[t]
    \centering
    \includegraphics[width=1\columnwidth]{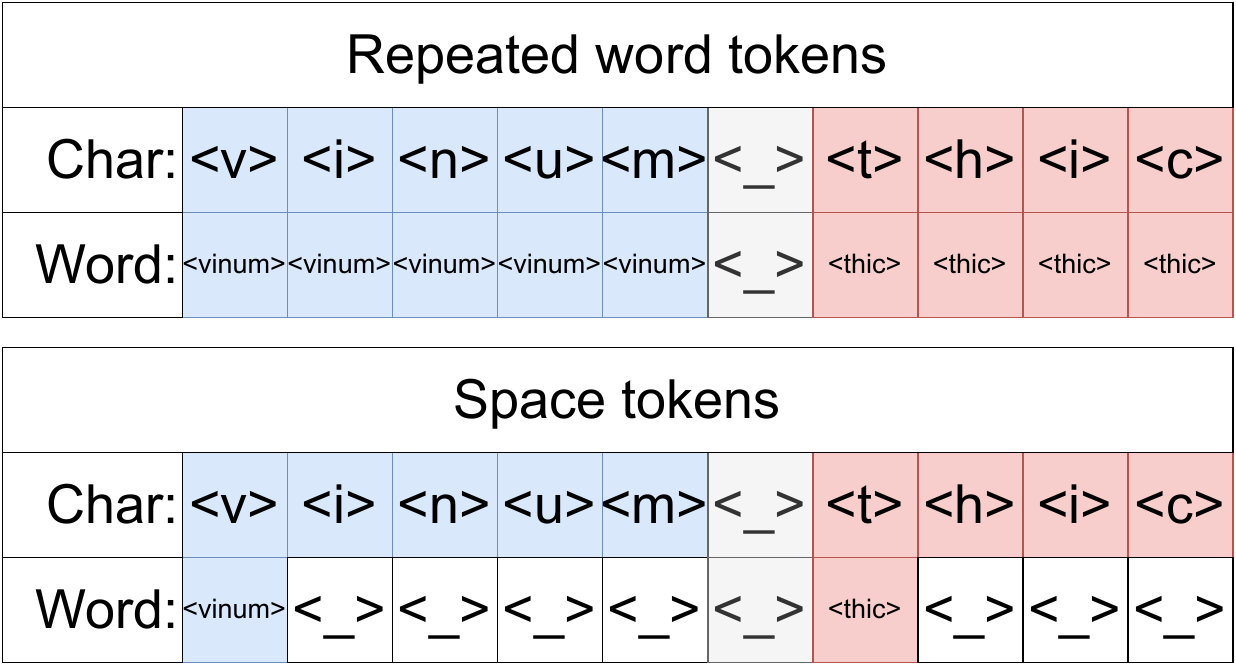}
    \caption{Example of how the character and word sequence are aligned. The string \textit{vinum thic} means \textit{wine and water}.}
    \label{fig:alignment}
\end{figure}

Finally, we propose a transformer model, \textbf{Larth}.
The encoder is based on Ithaca \citep{ithaca}. 
It takes both the characters and the words as input and concatenates their embeddings. 
Then, the sequence is encoded with a BigBird attention block \citep{bigbird}.
The character and word sequences are aligned so that they have the same length.
To do so, we test two approaches: we either extend the word sequence by repeating the word tokens or by adding space tokens as shown in Figure \ref{fig:alignment}.

\begin{figure}[t]
    \centering
    \includegraphics[width=0.9\columnwidth]{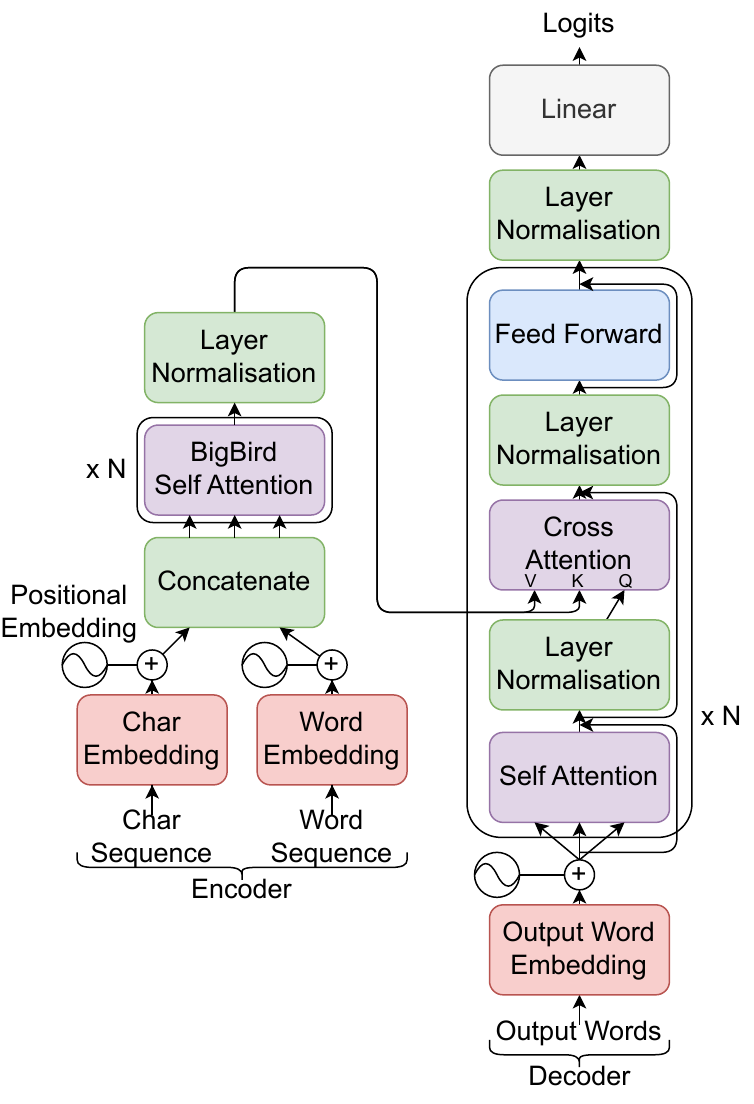}
    \caption{Transformer architecture used to translate Etruscan to English. The encoder imitates Ithaca's torso. For both the encoder and the decoder, one attention block is used.}
    \label{fig:transforemr}
\end{figure}

The decoder uses the encoded and the translated word sequences as input.
First, it applies self-attention to the translated sequence, and then it computes the cross-attention between the translation and the encoded inputs. 
A feed-forward layer generates the output.
Figure \ref{fig:transforemr} illustrates this architecture.

First, we train the model from scratch on Etruscan $\rightarrow$ English. 
Then, the model is initially trained for Latin $\rightarrow$ English or Ancient Greek $\rightarrow$ English and later fine-tuned on the original task Etruscan $\rightarrow$ English.

Moreover, we investigate the effect of using both the character and the word sequence by training with only one of the sequences and the effect of data augmentation
The model uses beam search when generating the output sequences, but
% We use one beam when evaluating during the training, and \note{§} beams when the training is completed.
we use one beam when evaluating during the training for efficiency.
Sequences are truncated at 256 tokens due to memory and computational resources.
\section{Experiments} \label{sec:experiments}

% \jnote{i would add a small paragraph here outlining what kind of experiments you are going to present so as the reader does not feel lost with all the subchapters.}

In this section, we compare different machine translation models trained on Etruscan data.
The models are compared on the BLEU score.
% In this section, we compare the BLEU scores of different machine translation models trained on Etruscan data.

\subsection{Random Model}
First, we run the random model on the Etruscan-Englih data. 
The dataset is split into 80 \% for training and 20 \% for testing. 
Only English labels are used for the training.
Each experiment is repeated 10 times with random dataset splits.
Table \ref{tab:random_model_results} reports the mean scores and the standard deviation of the models with different combinations of the datasets.

\begin{table}[t]
    \centering
    \begin{tabular}{c|ccc}
    \textbf{Dataset} & \textbf{BLEU} & \textbf{chr-F} & \textbf{TER} \\
    \hline
    ETP+ & 0.059 & 9.263 & 194.977 \\
    CIEP & (0.0174) & (0.295) & (10.676) \\
    \hline
    \multirow{2}*{ETP} & 0.324 & 13.970 & 133.878\\
    & (0.064) & (1.150) & (11.877) \\
    \end{tabular}
    \caption{Performance of the random model on the different Etruscan datasets. The table reports the mean value and the standard deviation of the metrics.}
    \label{tab:random_model_results}
\end{table}

\subsection{Dictionary-based Models}

From the book Zikh Rasna is possible to build a dictionary containing 821 vocables and their translations. 

We compare two tokenizers for Etruscan: 
% the first splits the tokens on the white spaces, 
the first uses white spaces to split the tokens,
and the second also separates the suffixes from the root.
The list of suffixes is also obtained from Zikh Rasna and the tokenizer recognises 178 suffixes.
Table \ref{tab:dictionary_results} shows the results of this model when translating Etruscan.

\begin{table}[t]
    \centering
    \begin{tabular}{c|ccc}
        \textbf{Dataset} & \textbf{BLEU} & \textbf{chr-F} & \textbf{TER} \\
        \hline
        ETP+CIEP & 0.167 & 9.120 & 89.799 \\
        ETP & 4.505 & 40.771 & 68.135 \\
        CIEP & 0.000 & 1.896 & 98.672 \\
        ETP (Suffix) & 1.605 & 37.669 & 82.666 \\
    \end{tabular}
    \caption{Results of the dictionary-based model when tested on the different sets. \textit{ETP (Suffix)} is the model tested on ETP with the suffix tokenizer.}
    \label{tab:dictionary_results}
\end{table}

If we consider the example \textit{"itun turuce venel atelinas tinas dlniiaras"} with the reference translation \textit{"venel atelinas dedicated this vase to the sons of tinia"}, this model predicts \textit{"this dedicated venel atelina tinia"}.
If we use the suffix tokenizer the prediction is 
\textit{"this for him dedicated three this venel laris atelina   shows"}.
\subsection{N-gram and Naive Bayes models}

Similarly to the random models, 80\% of the data is used for training, while the remaining 20\% is for testing. 
The dataset is ETP.
Each experiment is repeated 10 times with different random splits.

With the N-gram models, we compare models with a context size of 1, 2 and 3 that use only Etruscan or both Etruscan and English as context and whether they consider the word order.
Out-of-vocabulary (OOV) tokens are handled with additive smoothing.
We use 8 beams when generating the output sequence, however, this is equivalent to greedy search when the context uses only Etruscan.
Table \ref{tab:n-gram_results-ett} shows the results of the models that use only the Etruscan sequence, while Table \ref{tab:n-gram_results-etteng} shows the models that also use the English translations.

\begin{table}[t]
    \centering
    \begin{tabular}{c|ccc}
        \multicolumn{4}{c}{\textbf{Context: ETT - Word order: No}} \\
        \textbf{N-gram} & \textbf{BLEU} & \textbf{chr-F} & \textbf{TER} \\
        \hline
        \multirow{2}*{1} & \textbf{0.406} & \textbf{7.727} & \textbf{92.605} \\
         & (0.163) & (0.867) & (0.960) \\ 
        \multirow{2}*{2} & 0.006 & 3.249 & 98.035 \\
        & (0.001) & (0.752) & (0.821) \\
        \multirow{2}*{3} & 0.001 & 2.523 & 98.553 \\
        & (0.001) & (0.753) & (0.821) \\
        
        \multicolumn{4}{c}{\textbf{Context: ETT - Word order: Yes}} \\
        \textbf{N-gram} & \textbf{BLEU} & \textbf{chr-F} & \textbf{TER} \\
        \hline
        \multirow{2}*{1} & \textbf{0.405} & \textbf{7.727} & \textbf{92.605} \\
        & (0.163) & (0.867) & (0.960) \\
        \multirow{2}*{2} & 0.005 & 3.211 & 98.013 \\
        & (0.005) & (1.004) & (1.089) \\
        \multirow{2}*{3} & 0.001 & 2.523 & 98.531 \\
        & (0.001) & (0.748) & (0.870) \\
    \end{tabular}
    \caption{Mean scores and their standard deviation (in parenthesis) of the n-gram models that use only the Etruscan texts.}
    \label{tab:n-gram_results-ett}
\end{table}
\begin{table}[t]
    \centering
    \begin{tabular}{c|ccc}
        \multicolumn{4}{c}{\textbf{Context: ETT-ENG - Word order: No}} \\
        \textbf{N-gram} & \textbf{BLEU} & \textbf{chr-F} & \textbf{TER} \\
        \hline
        \multirow{2}*{1} & \textbf{0.218} & \textbf{3.059} & \textbf{92.902} \\
        & (0.018) & (0.301) & (1.160) \\
        \multirow{2}*{2} & 0 & 0 & 100 \\
        & (0) & (0) & (0) \\
        \multirow{2}*{3} & 0 & 0 & 100 \\
        & (0) & (0) & (0) \\
        
        \multicolumn{4}{c}{\textbf{Context: ETT-ENG - Word order: Yes}} \\
        \textbf{N-gram} & \textbf{BLEU} & \textbf{chr-F} & \textbf{TER} \\
        \hline
        \multirow{2}*{1} & \textbf{0.447} & \textbf{5.360} & \textbf{92.105} \\
        & (0.211) & (0.856) & (1.117) \\
        \multirow{2}*{2} & 0.000 & 0.370 & 99.705 \\
        & (0.000) & (0.167) & (0.346) \\
        \multirow{2}*{3} & 0.000 & 0.357 & 99.690 \\
        & (0.000) & (0.097) & (0.297) \\
    \end{tabular}
    \caption{Mean scores and their standard deviation (in parenthesis) of the n-gram models that use the Etruscan texts and the English translations. When the scores are zero is because the models immediately predict the end-of-sequence (EOS) token.}
    \label{tab:n-gram_results-etteng}
\end{table}

For the Naive Bayes models, we only use a context size of 2 and 3, and the models always consider the word order.
Table \ref{tab:naive_bayes_results} reports the results.

\begin{table}[t]
    \centering
    \begin{tabular}{cc|ccc}        
        \textbf{N} & \textbf{Context} & \textbf{BLEU} & \textbf{chr-F} & \textbf{TER} \\
        \hline
        \multirow{2}*{2} & \multirow{2}*{Ett.} & \textbf{0.160} & 12.609 & \textbf{101.482} \\
        & & (0.023) & (1.009) & (1.251) \\
        \multirow{2}*{3} & \multirow{2}*{Ett.} & 0.146 & \textbf{12.708} & 103.867 \\
        & & (0.030) & (0.921) & (1.220) \\
        \hline
        \multirow{2}*{2} & \multirow{2}*{Ett.-Eng.} & \textbf{0.055} & 9.547 & \textbf{101.522} \\
        & & (0.048) & (1.821) & (0.851) \\
        \multirow{2}*{3} & \multirow{2}*{Ett.-Eng.} & \textbf{0.055} & \textbf{9.954} & 103.038 \\
        & & (0.048) & (2.103) & (1.005) \\        
    \end{tabular}
    \caption{Mean scores and their standard deviation (in parenthesis) of the Na\"ive Bayes models.}
    \label{tab:naive_bayes_results}
\end{table}

\subsection{IBM models}
We split 80 \% of the data for training and 20 \%  for testing. 
Moreover, we use the previously built dictionary as training data.
No alignment information is given to the model, but IBM4 and IBM5 receive a dictionary that maps words to POS tags.
We assume that words can only have one tag.

IBM3, IBM4, and IBM5 are trained only with the dictionary data.
Models trained on ETP+CIEP are tested on ETP+CIEP, while models trained on ETP are tested on ETP as shown in Tables \ref{tab:ibm-etpciep} and \ref{tab:ibm-etp}.

\begin{table}[t]
    \centering
    \begin{tabular}{c|ccc}
        \multicolumn{4}{c}{\textbf{ETP+CIEP}}\\
        \textbf{Model} & \textbf{BLEU} & \textbf{chr-F} & \textbf{TER} \\
        \hline
        \multirow{2}*{IBM1} & \textbf{0.402} & \textbf{19.744} & \textbf{89.213} \\
         & (0.183) & (1.178) & (0.693) \\
        \multirow{2}*{IBM2} &  0.392 & 19.450 & 89.551 \\
         & (0.183) & (1.383) & (0.487) \\
        \multirow{2}*{IBM3(*)} & 0.105 & 8.629 & 91.052 \\
         & (0.046) & (1.148) & (1.507) \\
        \multirow{2}*{IBM4(*)} & 0.105 & 8.627 & 91.052 \\
         & (0.046) & (1.148) & (1.507) \\
        \multirow{2}*{IBM5(*)} & 0.105 & 8.631 & 91.063 \\
         & (0.046) & (1.147) & (1.516) \\
    \end{tabular}    
    \caption{Performance of the IBM models on the ETP+CIEP dataset. (*): IBM3, IBM4 and IBM5 are trained only with the dictionary.}
    \label{tab:ibm-etpciep}
\end{table}
\begin{table}[t]
    \centering
    \begin{tabular}{c|ccc}
        \multicolumn{4}{c}{\textbf{ETP}}\\
        \textbf{Model} & \textbf{BLEU} & \textbf{chr-F} & \textbf{TER} \\
        \hline
        \multirow{2}*{IBM1} & 2.187 & 37.363 & 73.917 \\
        & (0.596) & (2.011) & (2.163) \\
        \multirow{2}*{IBM2} & 2.104 & 36.721 & 74.334 \\
        & (0.449) & (2.098) & (2.090) \\
        \multirow{2}*{IBM3(*)} & \textbf{2.482} & \textbf{39.393} & \textbf{71.270} \\
        & (0.513) & (2.229) & (2.456) \\
        \multirow{2}*{IBM4(*)} & \textbf{2.482} & 39.391 & \textbf{71.270} \\
        & (0.514) & (2.228) & (2.456) \\
        \multirow{2}*{IBM5(*)} & 2.481 & 39.416 & 71.331 \\
        & (0.513) & (2.235) & (2.415) \\
    \end{tabular}
    \caption{Performance of the IBM models on the ETP dataset. (*): IBM3, IBM4 and IBM5 are trained only with the dictionary.}
    \label{tab:ibm-etp}
\end{table}

As an example, IBM3 translates \textit{"eca shuthic velus ezpus clensi cerine"} as \textit{"this funerary vel etspus son constructed"}, while the reference translation is \textit{"this funerary monument belongs to vel etspu it is constructed by his son"}.

\subsection{Transformer Models - Larth}

The model is trained for Etruscan $\rightarrow$ English translation with ETP+CIEP and with ETP only. 
The models are tested on the same split of the dataset.
Due to the small size of the dataset, 95 \% of the data is used for training.

The optimizer is RAdam \citep{radam}, with an initial learning rate of 0.002 and 250 warmup steps. 
We use a reverse square root learning schedule.
The loss function is cross-entropy, and the batch size is 32.
We set the label smoothing to 0.1.

We first try to train from scratch and with different alignment techniques. The BLEU, chr-F and TER scores are shown in Table \ref{tab:larth-et}.
We use data augmentation with ETP+CIEP with the sequences aligned by repeating the word tokens, however, we do not use it on ETP due to the decrease in performance.

% Data augmentation
\begin{table}[t]
    \centering    
    \begin{tabular}{c|ccc}
        \multicolumn{4}{c}{\textbf{ETP+CIEP}}\\
        \textbf{Model} & \textbf{BLEU} & \textbf{chr-F} & \textbf{TER} \\
        \hline
        repeat & \textbf{10.1} & 15.11 & \textbf{144.5} \\
        space & 5.201 & \textbf{16.9} & 274.8 \\
        repeat+unk & 2.8 & 14.8 & 189.1 \\
        repeat+name & 1.004 & 12.2 & 615.9 \\
        % ETP+CIEP+unk+name & repeat & - & - & - \\        
        \multicolumn{4}{c}{\textbf{ETP}}\\
        \textbf{Model} & \textbf{BLEU} & \textbf{chr-F} & \textbf{TER} \\
        \hline
        repeat & \textbf{9.053} & \textbf{17.24} & 137 \\
        space & 5.784 & 15.88 & \textbf{124.7} \\
        % ETP+unk & repeat & - & - & - \\
        % ETP+name & space & - & - & - \\
        % ETP+unk+name & repeat & - & - & - \\
    \end{tabular}
    \caption{Larth trained from scratch for Etruscan $\rightarrow$ English. \textit{Repeat} and \textit{space} indicate how the character and the word sequence are aligned. \textit{+name} is trained with data augmented by changing names, while \textit{+unk} is augmented by deleting characters.}
    % The models use both characters and words as input.}
    \label{tab:larth-et}
\end{table}

Next, we train the same architecture with only the word sequence or only the character sequence.
The results are shown in Table \ref{tab:larth-seqs}.

\begin{table}[t]
    \centering
    \begin{tabular}{c|ccc}
        \multicolumn{4}{c}{\textbf{ETP+CIEP}} \\
        \textbf{Inputs} & \textbf{BLEU} & \textbf{chr-F} & \textbf{TER} \\
        \hline
        char & 0.9694 & 14.42 & 254.8 \\
        word & 2.776 & 13.49 & \textbf{99.88} \\
        char+word & \textbf{10.1} & \textbf{15.11} & 144.5 \\
        \multicolumn{4}{c}{\textbf{ETP}} \\
        \textbf{Inputs} & \textbf{BLEU} & \textbf{chr-F} & \textbf{TER} \\
        \hline
        char & 0.1431 & 11.22 & 528.1 \\
        word & 7.679 & \textbf{18.48} & \textbf{131.6} \\
        char+word & \textbf{9.053} & 17.24 & 137 \\
    \end{tabular}
    \caption{Larth trained from scratch for Etruscan $\rightarrow$ English with only the character or the word sequence or both as input.}
    \label{tab:larth-seqs}
\end{table}

When training the same model with the Latin and Greek data, it achieved, respectively, BLEU/chr-F/TER of 0.4968/5.01/151.4 and 0.12/6.186/107.3.
Then, we fine-tune those models with Etruscan as shown in Table \ref{tab:larth-lat-grc}.

\begin{table}[t]
    \centering
    % \begin{tabular}{|c|c|c|c|}
    \begin{tabular}{c|ccc}
        % \hline
        \textbf{Data} & \textbf{BLEU} & \textbf{chr-F} & \textbf{TER} \\
        \hline%\hline
        Lat+ETP+CIEP & 0.1965 & 2.195 & 351.6 \\
        Grc+ETP+CIEP & \textbf{1.011} & \textbf{8.148} & \textbf{215.3} \\
        \hline
        Lat+ETP & 0.293 & 3.784 & 654.4 \\
        Grc+ETP & \textbf{2.037} & \textbf{6.04} & \textbf{164} \\
        % \hline
    \end{tabular}
    \caption{Larth trained with Latin (Lat) or Ancient Greek (Grc) and then fine-tuned on Etruscan.}
    \label{tab:larth-lat-grc}
\end{table}

Larth trained on ETP translates \textit{"mi aveles metienas"} as \textit{"i am the tomb"} while the reference translation is \textit{"i am the tomb of avele metienas"}. Note that in this example \textit{"the tomb"} is implied and not mentioned explicitly.

When trained on ETP+CIEP, we have \textit{"e ca shuthi anes cuclnies"} translated as \textit{"this tomb"} but the reference is \textit{"this is the tomb of ane cuclnies"}. In this case \textit{"the tomb"} is mentioned, but the model misses the name of the owner, which is also mentioned.

\section{Results \& Discussion} \label{sec:results}
\begin{figure}[t]
    \centering
    \includegraphics[width=1\columnwidth]{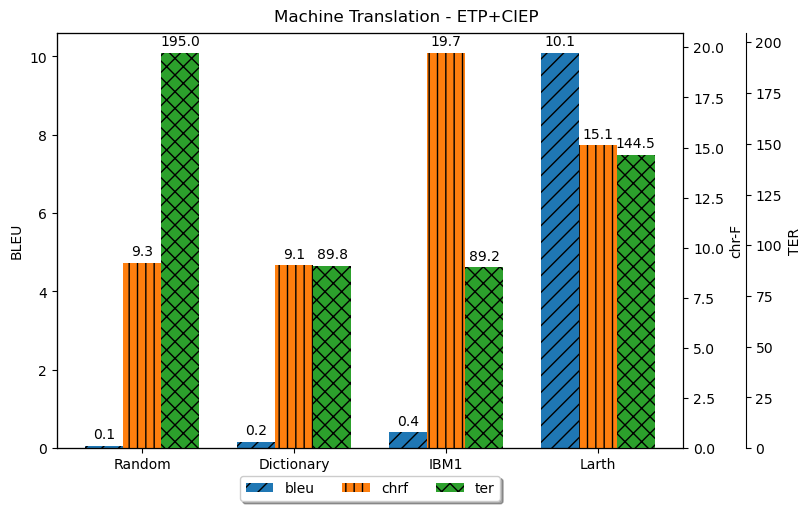}    
    \caption{Comparison of the models with the best BLEU scores on ETP+CIEP. One model from each type is selected.}
    \label{fig:comparison-etpciep}
\end{figure}

\begin{figure}[t]
    \centering
    \includegraphics[width=1\columnwidth]{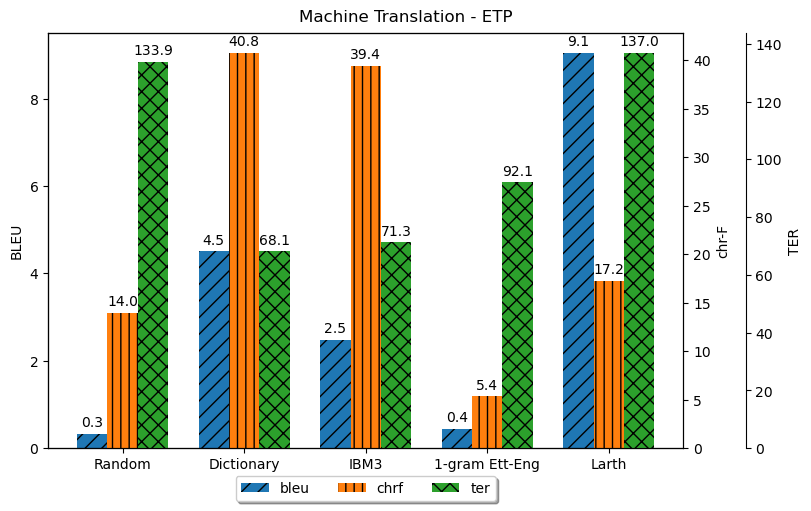}    
    \caption{Comparison of the models with the best BLEU scores on ETP. One model from each type is selected.}
    \label{fig:comparison-etp}
\end{figure}

%In the previous section, several models have been presented.
Figure \ref{fig:comparison-etpciep} and Figure \ref{fig:comparison-etp} compare the scores of the models presented in the previous Section. Compared to the random model, the dictionary-based model shows higher BLEU and chr-F scores and lower TER scores except when tested only on CIEP.
This suggests that CIEP is noisier than ETP and that the dictionary is not suited for CIEP.

The N-gram models perform better than random only when using unigrams as context. 
With longer n-grams, the performance decrease until the model only predicts the EOS token.
We can make similar observations for Na\"ive Bayes models.

IBM models are able to perform better than random.
When trained on ETP+CIEP, simpler models work better. 
This, again, might depend on the noise in CIEP.
IBM3 works better on ETP despite being trained only with the dictionary. 
Adding POS information (IBM4 and IBM5) does not improve the results.
However, on ETP the dictionary-based model still performs better than the IBM models. 

Larth is able to achieve a better BLEU score than the previous models on both ETP and ETP+CIEP.
However, it needs to use both the character and word sequences and the word tokens are repeated to align the two sequences, whereas the other models only use the word tokens.
Using the space token to align the sequences decrease the performance, but the BLEU score is still higher than the dictionary-based model.
A similar observation can be made for the model using only the word sequence. 
Using data augmentation or only the character sequences reduces the performance that is still higher than random.

Fine-tuning from Latin and Ancient Greek always performs worse than the dictionary-based model. 
This may depend on the small size of the model that is not able to adapt.
% \todo{Claim without support}

As for chr-F and TER, the dictionary model and IBM models perform better than Larth.
These two models can only output tokens from the training set and ignore unknown tokens. Thus, they can generate longer sequences of correct characters (high chr-F) and the errors are mainly for unknown tokens or from English tokens that are not directly present in the Etruscan texts like articles (low TER). 
Whereas, Larth uses tokens that can be word pieces and it still generates a translation for unknown tokens. 

\section{Discussion} \label{sec:conclusion}

In this paper, we present a dataset for Etruscan $\rightarrow$ English machine translation.
Although the dataset is not very big, we show that it is possible to train statistical and transformer models. Given the unexplored nature of Etruscan language, the fact that trained models perform better than random is an important first step for this language.
Moreoever, we demonstrated that Larth performs better than the IBM models when trained on the available data. 

However, our model does not provide any explanation about the generated translation neither it guarantees whether it is correct.
Our model's performance also depends on the dataset itself, which does not contain any bibliographic information or the reasoning that the original authors used to translate the inscriptions.
Future work includes delivering a cleaner and more complete version of the dataset and the inclusion of additional metadata, such as bibliographic information, more accurate location, or interesting graphical information (e.g. the direction of the inscription).%, the particular shape of some letters).

\clearpage
\bibliography{ref}
\bibliographystyle{acl_natbib}

\end{document}